\documentclass[letterpaper, 10 pt, conference]{ieeeconf}  

\IEEEoverridecommandlockouts                              

\usepackage{graphicx} 
\usepackage{amsmath} 
\usepackage{amssymb}  

\title{\LARGE \bf
Multiple-Lap Path Tracking for an Autonomous Race Vehicle via Iterative Learning Control }

\author{Nitin R. Kapania $^{1,2}$ and J. Christian Gerdes$^{1}$
\thanks{$^{1}$Department of Mechanical Engineering, Stanford University, Stanford, CA 94305, USA}
\thanks{$^{2}$Corresponding author: {\tt\small nkapania@stanford.edu}}
}

\begin{document}

\maketitle
\thispagestyle{empty}
\pagestyle{empty}

\begin{abstract}

Iterative learning control has been successfully used for several decades to improve the performance of control 
systems that perform a single repeated task. Using information from prior control executions, learning controllers gradually 
determine open-loop control inputs whose reference tracking performance can exceed that of traditional feedback-feedforward control algorithms. This paper considers
iterative learning control for a previously unexplored field: autonomous racing. Racecars are driven multiple laps around
the same sequence of turns while operating near the physical limits of tire-road friction, where steering dynamics become highly nonlinear and transient, making
accurate path tracking difficult. However, because the vehicle trajectory is identical for each lap in the case of single-car racing, the nonlinear vehicle dynamics
 and unmodelled road conditions are repeatable and can be accounted for using iterative learning control, provided the tire force limits have not been exceeded.
 This paper describes the design and application of proportional-derivative (PD) and quadratically
optimal (Q-ILC) learning algorithms for multiple-lap path tracking of an autonomous race vehicle. Simulation results are used to tune controller gains and test convergence, and 
experimental results are presented on an Audi TTS race vehicle driving several laps around Thunderhill Raceway in Willows, CA at lateral accelerations of up to 8 $\mathrm{m/s^2}$. Both
control algorithms are able to correct transient path tracking errors and improve the performance provided by a reference feedforward controller.

\end{abstract}

\section{Introduction}

Iterative learning control (ILC) solves the control problem of tracking a specific reference command that is repeated many times. 
By using tracking error information from prior attempts, ILC techniques are used to gradually determine the open-loop
control inputs that cause the system output to track a desired reference with minimal tracking error. 

ILC algorithms work best when the exogenous system signals (i.e. disturbances and reference inputs)
are constant from iteration-to-iteration. As a result, iterative learning controllers have frequently been applied in relatively structured robotics and automation environments, with recent publications 
considering piezolectric positioning \cite{huang}, robotic arm tracking \cite{freeman},  and microdeposition \cite{hoelzle}. However, 
iterative learning control has recently expanded into applications
outside of the traditional automation and process control setting. Chen and Moore \cite{chen} proposed a simple iterative learning scheme in 2006 to improve path-following of a ground vehicle with omni-directional wheels, where
 double integration of the previous feedback input was used to improve the feedforward signal. In 2013, Sun et al. proposed an iterative learning controller for overspeed proctection of high-speed
trains \cite{sun}. Purwin and Andrea synthesized an iterative controller using least-squares methods to aggressively maneuver a quadrotor unmanned aerial vehicle from
one state to another \cite{purwin}. Iterative learning control has also extended into the neuromuscular and biological domains, with Rogers et al. \cite{rogers} developing an ILC algorithm 
for robotically assisted stroke rehabilitation.

This paper presents autonomous race car piloting as a previously unexplored application area for iterative learning control. Race car drivers must drive multiple laps around
the same sequence of turns on a closed track, while operating near the physical limits of tire-road friction to minimize lap times.
 At the limits of tire friction, the steering dynamics associated with lateral  tracking of the desired path become highly nonlinear,
and difficult-to-measure disturbances such as bank, grade and local friction variation of the road surface have a large effect on the transient dynamics of the vehicle. 
These factors make development of a suitable feedforward steering controller challenging. However, because operation of the race vehicle 
occurs over multiple laps, with the reference road curvature unchanging from lap-to-lap, the unknown transient disturbances and
 vehicle dynamics tend be constant from lap to lap and can therefore be accounted for via iterative learning control. A notable exception occurs when the vehicle has significantly
 understeered due to lack of front tire force availability. In this case, additional steering will have no impact on path tracking.
 
 This paper is further divided as follows. Section II introduces a linear model for the planar vehicle dynamics of a race car following a fixed reference path. Because the transfer function between
the steering wheel input and the vehicle's path deviation is open-loop unstable, a stabilizing lanekeeping controller is added to the steering system and the closed loop dynamics
are represented in the commonly used ``lifted domain". Section III presents a PD-type iterative learning controller with a low-pass filter used to speed
up convergence. Gain tuning and stability at low lateral accelerations are shown using lifted domain techniques, while nonlinear simulations are used to predict a desirable tracking response in the presence of
high lateral acceleration. Section IV presents a quadratically optimal (Q-ILC) iterative learning controller, which has the benefit of explicitly
accounting for changes in vehicle speed along the race track. Section V presents experimental data of both controllers implemented on an Audi TTS race vehicle at combined 
lateral/longitudinal accelerations of up to 8 $\mathrm{m/s^2}$.
 
\section{Vehicle Dynamics and Problem Overview}
\label{sec:problemDescription}

\begin{figure}[h]
\centering
\includegraphics[width=3in]{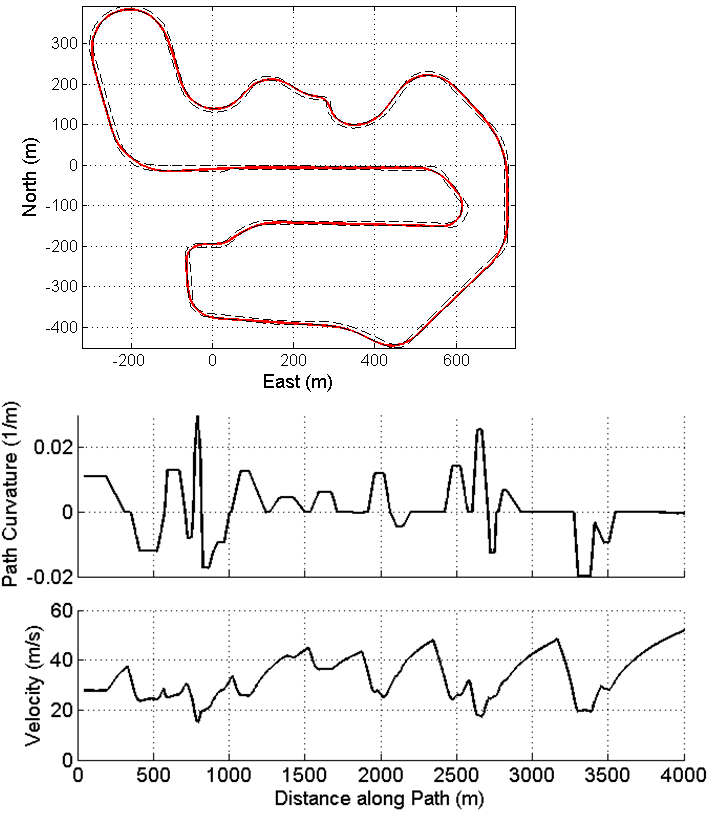}
\caption{(a) Overhead plot of racing line for Thunderhill Raceway, Willows CA, USA. (b) Racing line represented by curvature along distance traveled. (c) Velocity profile
to keep vehicle at combined lateral/longitudinal acceleration of 8 $\mathrm{m/s^2}$.}
\label{problemInfo}
\end{figure}

For this paper, the objective of piloting an autonomous vehicle along a fixed race track in minimum time is divided into separate lateral and longitudinal vehicle control problems. The lateral 
controller uses the the steering wheel input to track a desired ``racing-line", shown in 
Fig.~\ref{problemInfo}a. The racing line is frequently represented by a path curvature function $\kappa(s)$ parametrized by distance along the track (Fig. \ref{problemInfo})b. 
The longitudinal controller tracks a desired speed profile that keeps the vehicle at a specified lateral-longitudinal acceleration magnitude, typically near the limits of tire-road friction (Fig.~\ref{problemInfo})c. 
This paper focuses on learning the desired steer angle command $\delta_L$ over multiple laps in order to accurately track the reference path at high speeds, and we therefore assume minimum time 
velocity and curvature profiles have been computed using methods published in \cite{theodosis}, and that tracking of the velocity profile is handled by a separate controller \cite{mickthesis}. 

\subsection{Lateral Vehicle Dynamics}

Fig.~\ref{fig:bikemodel} shows a schematic of a vehicle following a path
with curvature profile $\kappa(s)$. The lateral deviation of the vehicle from the desired path ($e$) is the measured output for the ILC algorithms.
Since iterative learning controllers determine appropriate \textit{feedforward} control inputs over several iterations, it is necessary for the open-loop dynamics between the control input and
control output to be asymptotally stable. For the case of active steering control, the transfer function between the vehicle steer angle ($\delta$) and the lateral path error $e$ is
characterized by two poles at the origin, requiring the addition of a stabilizing feedback controller.

A lookahead controller provides the stabilizing feedback command for this paper. The ``lookahead error" is defined by 

\begin{equation}
	e_\mathrm{LA}=e+x_\mathrm{LA}\Delta\Psi
\end{equation}

Where $\Delta\Psi$ is the vehicle heading error and $x_\mathrm{LA}$ is the lookahead distance, typically 5-20 meters for autonomous driving.  The resulting feedback control law is

\begin{equation}
	\delta_\mathrm{FB} = -k_\mathrm{P}e_\mathrm{LA}
	\label{eqn:lookahead}
\end{equation}

with proportional gain $k_\mathrm{P}$. The control law (\ref{eqn:lookahead}) is a natural extension of potential field lanekeeping, as described by Rossetter et al. in \cite{rossetter2002}, which also provides heuristics for 
selecting $k_\mathrm{P}$ and $x_\mathrm{LA}$. Desirable stability properties over significant tire saturation levels are demonstrated in \cite{talvala}.

With the feedback controller added, closed loop dynamics of the lateral path deviation are dependent on three other states: vehicle sideslip $\beta$, yaw rate $r$ and heading error $\Delta\Psi$. For controller development
and testing, these dynamics are given by the planar bicycle model:

\begin{figure}
\centering
\includegraphics[width=3.5in]{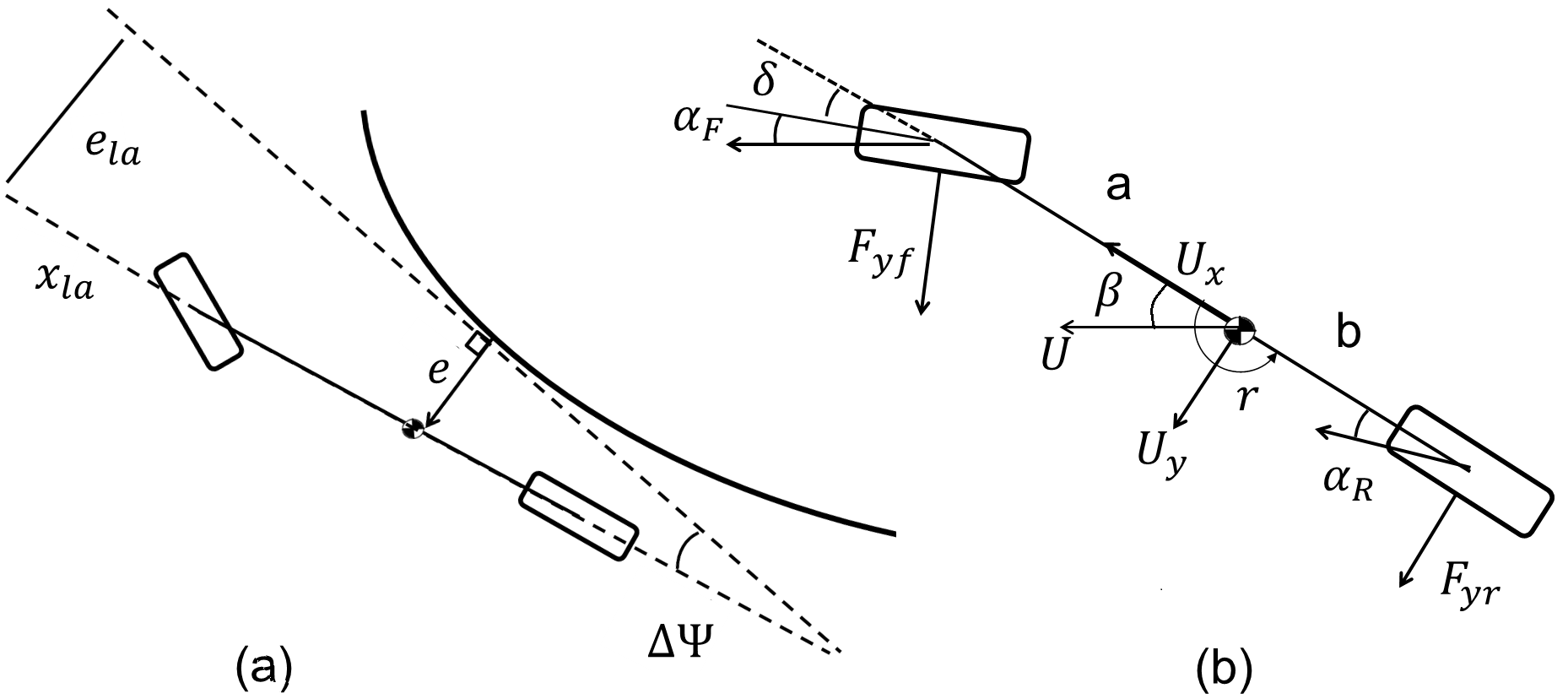}
\caption{Left: Schematic of bicycle model. Right: Diagram showing tracking error $e$, heading error $\Delta\Psi$, lookahead distance $x_\mathrm{LA}$, and lookahead error $e_\mathrm{LA}$.}
\label{fig:bikemodel}
\end{figure}

\begin{subequations}
\label{eq:bm}
\begin{align}
	\dot{\beta} &= \frac{F_\mathrm{yf}+F_\mathrm{yr}}{mU_x} - r \qquad \dot{r} = \frac{aF_\mathrm{yf} - bF_\mathrm{yr}}{I_z} \label{bm1} \\
	\dot{e} &= U_x (\beta + \Delta\Psi) \qquad \Delta\dot{\Psi} = r - U_x\kappa \label{eq:bm2} 
\end{align}
\end{subequations}

Where $U_x$ is the vehicle forward velocity and $F_{yf}$ and $F_{yr}$ the front and rear lateral tire forces. The vehicle mass and yaw inertia are denoted by $m$ and $I_z$, while the geometric
parameters $a$ and $b$ are shown in Fig.~\ref{fig:bikemodel}.

As automotive racing frequently occurs near the limits of tire force saturation, 
lateral tire force is modeled using the nonlinear Fiala brush tire model, assuming a single coefficient of friction and a parabolic force distribution \cite{Pacejka2012}.
 The lateral tire forces are functions of the front and rear tire slip angles $\alpha_\mathrm{F}$ and $\alpha_\mathrm{R}$. 

\begin{eqnarray}
\label{eq:fiala}
\small
	F_\mathrm{y}&=&\begin{cases} -C_{\alpha}\tan\alpha + \frac{C_{\alpha}^2}{3\mu F_\mathrm{z}} |\tan\alpha| \tan\alpha \vspace{2mm} \\ \hspace{4mm}- \frac{C_{\alpha}^3}{27\mu^2F_\mathrm{z}^2}\tan^3\alpha,
\hspace{4mm}  |\alpha| < \arctan{\left(\frac{3\mu F_\mathrm{z}}{C_\alpha}\right)} \\ -\mu F_\mathrm{z}\text{sgn} \ \alpha, \hspace{14mm} \mathrm{otherwise} \end{cases} \nonumber\\
\end{eqnarray}

where $\mu$ is the surface coefficient of friction, $F_\mathrm{z}$ is the normal load, and $C_\alpha$ is the tire cornering stiffness. The linearized tire slip angles are given by

\begin{subequations}
\begin{align}
	\alpha_\mathrm{F} &= \beta + \frac{ar}{U_\mathrm{x}} - \delta\\
	\alpha_\mathrm{R} &= \beta - \frac{br}{U_\mathrm{x}}
\end{align}
\end{subequations}

\subsection{Linear Time Varying Model in the Lifted Domain}

While the nonlinear tire model presented in (\ref{eq:fiala}) captures the effect of tire saturation at the limits of handling, established methods for design and analysis
of iterative learning controllers requires a linear system description. With the assumption of low lateral acceleration, a simple linear tire model is given by, 

\begin{align}
	F_y = -C_\alpha\alpha
\end{align}

The lateral error dynamics of the vehicle in response to a \textit{feedforward} steer angle input (recall that there is also a feedback steer angle $\delta_\mathrm{FB}$ in the loop)
 can then be represented by the continuous, linear time varying (LTV) state space model,

\begin{subequations}
\label{eq:cts}
\begin{align}
	\dot{x}(t) &= A_c(t) x + B_c(t)\delta_L + d_c(t)\\
	 e &= C_cx(t)\\
	 x &= [e \hspace{2mm} \Delta\Psi \hspace{2mm} r \hspace{2mm} \beta]^T
\end{align}
\end{subequations}
Where $\delta_L$ is the learned steering input. The LTV framework is chosen to account for the variation in vehicle 
velocity $U_x$ along the track (Fig.~\ref{problemInfo}). While the longitudinal dynamics are not explicitly accounted for in (\ref{bm1}), 
allowing for the state matrices to vary with measured values of $U_x$ enables more accurate controllers and analysis. The time-varying matrices $A_c$, $B_c$, and $C_c$ are given by

\begin{multline}
\label{eqn:Amatrix}
A_c(t)  =  \\
\left[\begin{smallmatrix}
  0 & U_\mathrm{x}(t) & 0 & U_\mathrm{x}(t) \\ 
  0 & 0 & 1 & 0 \\ 
  \frac{-ak_\mathrm{P} C_\mathrm{F}}{I_\mathrm{z}}  & \frac{-ak_\mathrm{P}x_{LA}C_\mathrm{F}}{I_\mathrm{z}}  & \frac{-(a^2C_\mathrm{F}+b^2C_\mathrm{R})}{U_\mathrm{x}(t)I_\mathrm{z}} & \frac{bC_\mathrm{R} - aC_\mathrm{F}}{I_\mathrm{z}}  \\
  \frac{-k_\mathrm{P}C_\mathrm{F}}{mU_\mathrm{x}(t)}  & \frac{-k_\mathrm{P}x_{LA}C_\mathrm{F}}{mU_\mathrm{x}(t)}  & \frac{bC_\mathrm{r}-aC_\mathrm{f}}{mU_\mathrm{x}(t)^2}-1 & \frac{-(C_\mathrm{F} + C_\mathrm{R})}{mU_\mathrm{x}(t)}
 \end{smallmatrix}\right]
 \end{multline}

\begin{align}
B_c(t) =[0 \hspace{2 mm} 0 \hspace{3 mm} \frac{a C_\mathrm{F}}{I_\mathrm{z}} \hspace{3 mm}  \frac{C_\mathrm{F}}{mU_\mathrm{x}(t)}]^T
\end {align}
\begin{align}
C_c =[1 \hspace{2 mm} 0 \hspace{2 mm} 0 \hspace{2 mm} 0]
\end{align}

The disturbance $d_c$, is assumed constant from lap to lap and is given by

\begin{align}
d_c(t) = [0 \hspace{2 mm} -\kappa U_x(t) \hspace{3 mm} 0 \hspace{3 mm}  0]^T
\end{align}

The next step is to discretize (\ref{eq:cts}) by the controller sample
time $T_s$, resulting in the discrete time system,

\begin{align}
x_j(k\!\!+\!\!1) &= A(k)x_j(k) + B(k)\delta^L_j(k)\\
e_j(k) &= Cx_j(k)
\label{eq:discreteEOM}
\end{align}

Where $k = 1\ldots N$ is the time sample index, and $j = 1\ldots M$ is the number of iterations (i.e. the number of laps around the track). 
Development and analysis of the iterative learning controllers in the next section will be made easier by representing the system dynamics in the ``lifted-domain",
where the inputs  and outputs  are stacked into arrays and related by matrix multiplication, as follows:

\begin{subequations}
\label{eq:liftedDomain}
\begin{align}
\mathbf{e}_j        &= \mathbf{P}\mathbf{\delta^L_j} + \mathbf{d}_j\\
\bf{\delta}_j^L  &=[\delta^L_j(0) \cdots \delta^L_j(N\!\!-\!\!1)]^T\\
\bf{e}_j        &= [e_j(1) \cdots e_j(N)]^T
\end{align}
\end{subequations}

Where the elements of the $N \times  N$ matrix $\bf{P}$ are given by,

\begin{equation}
p_{lk} = \begin{cases} 0 &\mbox{if } l < k \\ 
CB(k) & \mbox{if } l = k \\
CA(l)A(l-1)\cdots A(k)B(k) &\mbox{if } l > k \end{cases}
\end{equation} 

Note that for the case where $U_x$ is constant, we have a linear time invariant (LTI) system given by only
$N$ elements, with $p(k) = CA^{k-1}B$, and the resulting lifted-domain matrix $\bf{P}$ is Toeplitz. 

\section{Controller Design}

With the representation of the steering dynamics for a given lap represented by (\ref{eq:liftedDomain}), the next step is to
design algorithms that determine the learned steering input $\bf{\delta}^L_{j\!+\!1}$ for the next lap, given the error response
from the completed lap $\bf{e_{j}}$. A common framework for iterative learning algorithms is to choose $\bf{\delta}^L_{j\!+\!1}$ as follows \cite{bristow},

\begin{align}
\bf{\delta}^L_{j\!+\!1} = Q(\bf{\delta_{L,j}} - L\bf{e_{j}})
\label{eq:ctrlLaw}
\end{align}

Where $\bf{Q}$ is the $N \times N$ filter matrix, and $\bf{L}$ is the $N \times N$ learning matrix. In following sections, the
matrices $\bf{Q}$ and $\bf{L}$ will be obtained by designing a PD type iterative learning controller as well as a quadratically
optimal (Q-ILC) learning controller.

\subsection{Proportional-Derivative Controller}

The proportional-derivative iterative learning controller computes the steering addition $\delta^L_j(k)$ at a given time index $k$ based on the error $e_{j\!-\!1}(k)$ and the change in error
at the same time index from the previous lap,

\begin{equation}
	\delta^L_{{j\!+\!1}}(k) = \delta^L_j(k) - k_pe_{j\!-\!1}(k) - k_d(e_{j\!-\!1}(k) - e_{j\!-\!1}(k-1))
\end{equation}

Where $k_p$ and $k_d$ are the proportional and derivative gains. In the lifted domain representation from (\ref{eq:ctrlLaw}), the resulting learning matrix $\bf{L}$ is given by
\begin{equation}
	\bf{L} = \begin{bmatrix} -(k_p\!+\!k_d) &               &    0 \\ 
						 k_d           &  \ddots  &   \\ 
						 0             &   k_d            &    -(k_p\!+\!k_d)  \\ \end{bmatrix}
\end{equation}
 
 An important consideration in choosing the gains $k_p$ and $k_d$ is achieving a monotonic decrease in the path tracking error on every lap. 
 This property, known as monotonic convergence, occurs if the following condition is met \cite{bristow}:

\begin{equation}
	\gamma \triangleq \bar{\sigma}(\mathbf{P}\mathbf{Q}(I-\mathbf{L}\mathbf{P})\mathbf{P}^{-1}) < 1
	\label{eq:MS}
\end{equation}
	
Where $\bar{\sigma}$ is the maximum singular value. In this case, the value of $\gamma$ provides an upper bound on the decrease in the tracking error norm from lap to lap, i.e. 
\begin{equation}
	||e_\infty - e_{j+1}||_2 \leq \gamma ||e_\infty-e_j||_2
\end{equation}
where $e_\infty$ is the converged path tracking error. The monotonic stability condition differs from its weaker counterpart of asymptotic stability in that we are guaranteed that 
 the path tracking error on the first lap (i.e. with no added learning input) is the worst case performance.

\begin{figure}
\centering
\includegraphics[width=3.5 in]{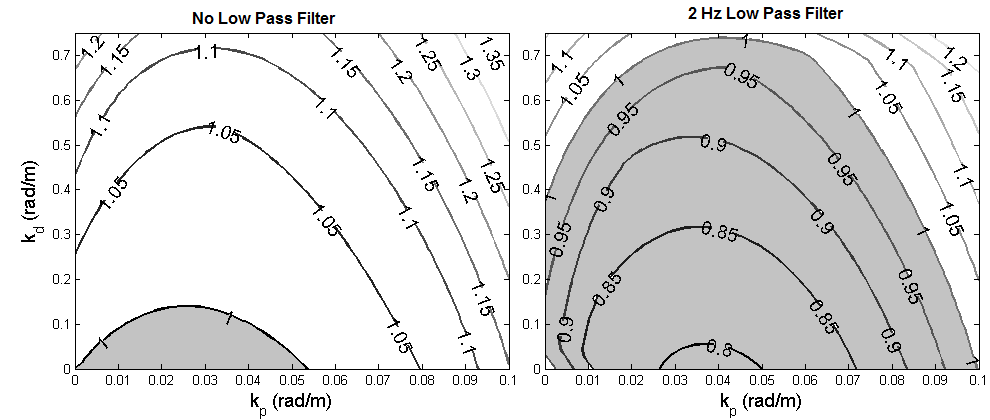}
\caption{Values of convergence bound $\gamma$ vs. $k_p$ and $k_d$ for PD iterative learning controller with (left) no filtering and (right) with a 2 Hz low-pass filter. Lower values of $\gamma$ correspond to faster convergence.
Shaded regions correspond to systems with monotonic stability. }
\label{fig:stabPlot}
\end{figure}

Fig. \ref{fig:stabPlot} shows values of $\gamma$ for both an unfiltered PD controller ($\mathbf{Q} = I$), and for a PD controller with a 2 Hz low pass filter. The $\gamma$ values are plotted as a
contour map against the controller gains $k_p$ and $k_d$. Addition of the low-pass filter
assists in achieving controller monotonic stability by removing oscillations that are frequently generated by iterative learning controllers when trying to remove small reference tracking
errors after several iterations. Since the filtering occurs when generating a control signal for the next lap, the filter $\mathbf{Q}$ can be zero-phase. 

\begin{figure}
\centering
\includegraphics[width=3 in]{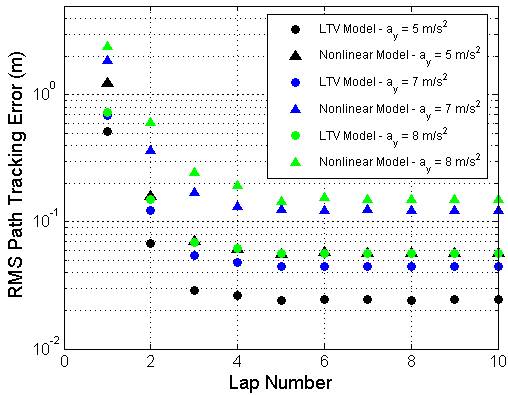}
\caption{Simulated results for root-mean-square path tracking error for PD iterative learning control at several values of vehicle lateral acceleration. Controller gains are $k_p = .05$, $k_d = .05$. Results from a nonlinear vehicle dynamics simulation are compared to results
from the linear vehicle model. }
\label{fig:PDsimResults}
\end{figure}

However, testing for linear stability is insufficient controller design given that racing frequently occurs near the limits of vehicle handling, when the vehicle dynamics are described 
by the nonlinear equations of motion presented in Section \ref{sec:problemDescription}. To test the PD controller feasibility, the vehicle tracking performance over multiple laps is
simulated using the path curvature and speed profile shown in Fig.~\ref{problemInfo}.
Simulated results for the root-mean-square (RMS) tracking error are shown in Fig.~\ref{fig:PDsimResults} for both the linear state dynamics prescribed by (\ref{eq:discreteEOM}) and 
the nonlinear dynamics model given by (\ref{eq:bm})
and (\ref{eq:fiala}). The results indicate that as the vehicle corners closer to the limits of handling, the tracking performance of the ILC degrades relative to the expected performance
given by the linear model, but can still be expected to converge over relatively few iterations. 

\subsection{Quadratically Optimal Controller}\label{sec:controller}

An alternate approach to determining the learned steering input $\bf{\delta}^L_j$ is to minimize a quadratic cost function for the next lap:

\begin{align}
J_{j\!+\!1} = e_{j\!+\!1}^TTe_{j\!+\!1} + \delta^{L\,T}_{j\!+\!1} R \; \delta^L_{j\!+\!1}+\Delta_{j\!+\!1}^TS\Delta_{j\!+\!1}
\label{eq:QILC}
\end{align}

Where $\Delta_{j\!+\!1} = \mathbf{\delta^L_{j\!+\!1}} - \mathbf{\delta^L_{j}}$ and the $N \times N$ matrices $T$, $R$, and $S$ are weighting matrices.
This formulation allows the control designer to weight the competing objectives of minimizing tracking error, control effort, and change in the control signal from lap to lap.
While constraints can be added to the optimization problem, the unconstrained problem in (\ref{eq:QILC}) can be solved analytically \cite{bristow} to obtain desired controller and filter matrices:

\begin{subequations}
\label{eq:analSol}
\begin{align}
	\mathbf{Q} &= (P^TTP + R + S)^{-1}(P^TTP+S)\\
	\mathbf{L} &= (P^TTP + S)^{-1}P^TT
\end{align}
\end{subequations}

An advantage to the quadratically optimal control design over the simple PD controller is that the controller matrices $\mathbf{Q}$ and $\mathbf{L}$ take the linear time-varying dynamics $\mathbf{P}$ into account. This allows the iterative learning algorithm
to take into account changes in the steering dynamics due to changes in vehicle velocity. However, a disadvantage is that determining new $\mathbf{Q}$ and $\mathbf{L}$ matrices every lap
requires resolving (\ref{eq:analSol}), which can be computationally expensive for fast sampling rates. 

\begin{figure}
\centering
\includegraphics[width=3 in]{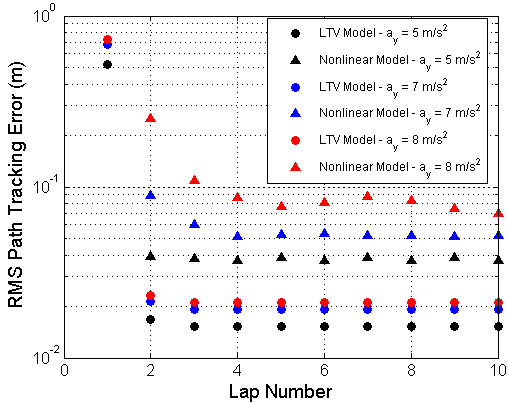}
\caption{Simulated results for root-mean-square path tracking error for Q-ILC at several values of vehicle lateral acceleration, with $T = R = I$ and $S = \mathrm{100} I$. Results from a nonlinear vehicle dynamics simulation are compared to results
from the linear vehicle model.}
\label{fig:QILCsimResults}
\end{figure}

Fig.~\ref{fig:QILCsimResults} shows simulated results for the quadratically optimal ILC controller. The results are similar to those in Fig.~\ref{fig:PDsimResults} in that the predicted controller performance
is worse when the simulation accounts for nonlinear tire dynamics. However, the simulation still shows a rapid decrease in path tracking error over the first ten laps, with RMS errors on the order
of 8-9 cm at lateral accelerations of .8 $g$.

\section{Experimental Results}

\begin{figure}
\centering
\includegraphics[width=3.5 in]{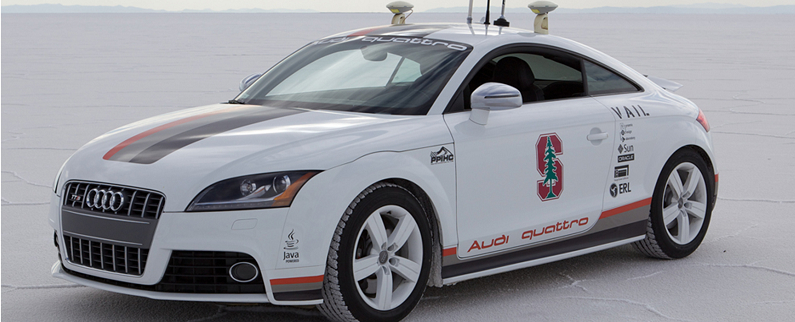}
\caption{Autonomous Audi TTS with electronic power steering, brake booster, and throttle by wire.}
\label{fig:shelleyPic}
\end{figure}

Experimental data for both controllers was collected over multiple laps at Thunderhill Raceway, a 3 mile paved racetrack in Willows, CA, with track boundaries shown in Fig.~\ref{problemInfo}a. The experimental testbed is an autonomous Audi TTS
equipped with an electronic power steering motor, active brake booster, and throttle by wire Fig.~\ref{fig:shelleyPic}. Vehicle
and controller parameters are shown in Table 1. 

An integrated Differential Global Positioning System (DGPS) and Inertial Measurement Unit (IMU) is used to obtain vehicle state information, and a localization algorithm 
determines the lateral path tracking error $e$, heading error $\Delta\Psi$, and distance along the desired racing line. The steering controller updates at 200 Hz, 
and the iterative steering corrections are calculated after each lap from data downsampled to 10 Hz. The corrections are then applied as a function of distance along the track using an
interpolated lookup table. For safety reasons, a steady-state feedforward steering algorithm \cite{kapania} is also applied to keep the tracking error on the first lap below 1 m.

\begin{table}[h]
\small
\begin{center}
\caption{Vehicle Parameters}\label{tb:params}
\begin{tabular}{lccc}
Parameter & Symbol & Value & Units \\\hline
Vehicle mass & $m$ & 1500 & kg \\
Yaw moment of inertia & $I_z$ & 2250 & $\mathrm{kg \cdot m}^2$\\
Front axle to CG & $a$ & 1.04 & m\\
Rear axle to CG & $b$ & 1.42 & m\\
Front cornering stiffness & $\mathrm{C}_\mathrm{F}$ & 160 & $\mathrm{kN \cdot rad}^{-1}$ \\
Rear cornering stiffness & $\mathrm{C}_\mathrm{R}$ & 180 & $\mathrm{kN \cdot rad}^{-1}$ \\
Lookahead Distance       & $x_\mathrm{LA}$          &  15.2 & $\mathrm{m} $ \\
Lanekeeping Gain         & $k_{\mathrm{LK}}$         & .053 & $\mathrm{rad\,m^{-1}}$\\
Lanekeeping Sample Time  & $t_s$                        & .005 & s\\
ILC Sample Time          & $T_s$                        & .1   & s\\
PD Gains                 & $k_p$ and $k_d$           & .02 \& .4 & $\mathrm{rad\,m^{-1}}$\\
Q-ILC Matrix             & $T$ and $R$              &  $I$      & - \\
Q-ILC Matrix             & $S$                       & 100\,$I$  & - \\\hline
\end{tabular}
\end{center}
\end{table}

\begin{figure}[h]
\centering
\includegraphics[width=3.5 in]{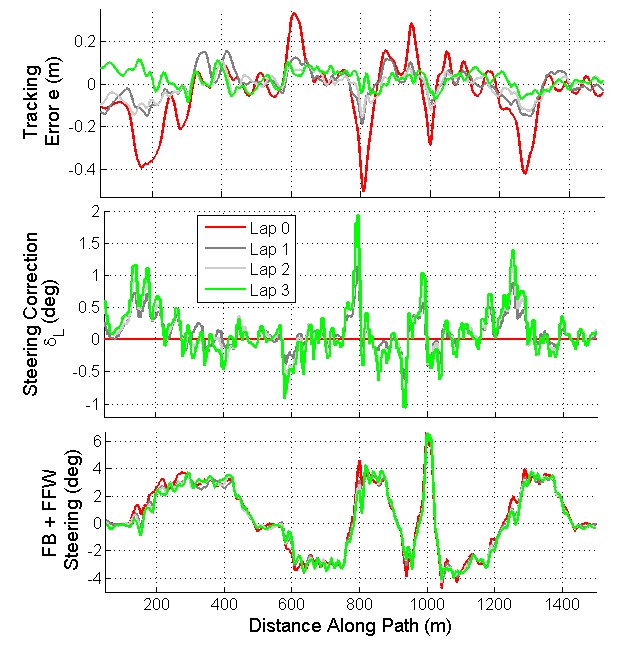}
\caption{Experimental results for path tracking error with PD learning controller, at peak lateral accelerations of 8 $m/s^2$.}
\label{fig:PDexp}
\end{figure}

Fig.~\ref{fig:PDexp} shows the applied iterative learning signals and resulting path tracking error over four laps using the PD learning algorithm. The car is driven
aggressively at peak lateral/longitudinal accelerations of 8 $\mathrm{m/s^2}$. On the first lap, despite the 
incorporation of a feedforward-feedback controller operating at a high sampling rate, several transient spikes in tracking error are visible due to the underdamped tire dynamics near the limits of handling.
However, the iterative learning algorithm is able to significantly attenuate these transient spikes over just two or three laps. Similar qualitative
results occur for the quadratically optimal ILC. 

\begin{figure}
\centering
\includegraphics[width=3 in]{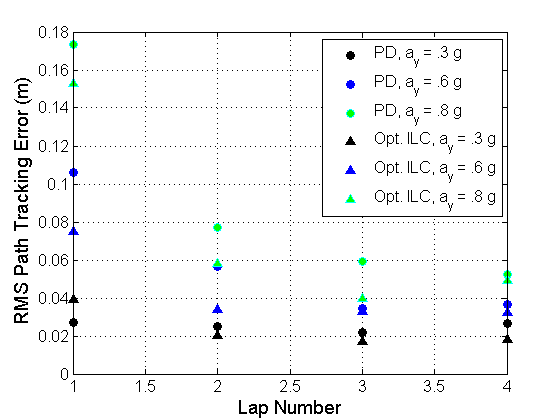}
\caption{Experimental results for both controller types over a variety of lateral accelerations.}
\label{fig:allExp}
\end{figure}

In Fig.~\ref{fig:allExp}, results are shown for versions of the velocity profile in Fig.~\ref{problemInfo} that are scaled to achieve different vehicle accelerations. The results
show that at low vehicle accelerations, when vehicle dynamics are accurately prescribed by a linear tire model, the feedback-feedforward controller is able to keep the vehicle close to the
desired path, leaving little room for improvement through iterative learning control. However, as the speed profile becomes more aggressive, the path tracking degrades in the presence
of highly transient tire dynamics, and iterative learning control can be effectively used to obtain tight path tracking path over two or three laps of racing. 
In practice, the performance of both the PD algorithm and quadratically optimal algorithms are similar, and an important observation is that the RMS tracking error increases slightly from
lap-to-lap at the end of some tests. While not predicted in simulation, this behavior is not unreasonable given unmodelled sensor noise and disturbances that vary from lap to lap. 
More refined tuning of the gain matrices may be able to prevent this RMS error increase, or the ILC algorithm can be stopped after several iterations once the tracking performance is acceptable.

\section{Conclusion}
This paper demonstrates the application of iterative learning control (ILC) methods to achieve accurate steering control
of an autonomous race car over multiple laps. Two different algorithms, proportional-derivative (PD) and 
quadratically optimal (Q-ILC) learning control are tested in simulation and then used to experimentally 
eliminate path tracking errors caused by the highly transient nature of the vehicle lateral dynamics near the limits of tire-road friction. 
Both learning algorithms provide comparable lap-to-lap tracking performance, although the PD method is computationally fast enough to run in real time. Two clear limitations of the presented iterative learning controllers present avenues for future work. While representing the vehicle dynamics
with a linear, time-varying model allows for the quadratically optimal ILC algorithm to account for varying longitudinal speed, a better approach is to linearize the vehicle
dynamics at each point on the track and create an affine time-varying model whose path tracking error can be minimized. Furthermore, applying a steering wheel input to eliminate
lateral errors will work only if the vehicle is near the limits of handling, but has not fully saturated the available tire force and entered a limit understeer condition.
In this case, a separate controller must be developed to modify the racing line and velocity profile for future laps in order to reduce the vehicle cornering forces.




\section*{ACKNOWLEDGMENT}
This research is supported by the National Science Foundation Graduate Research Fellowship Program (GRFP). 
The authors would like to thank the members of the Dynamic Design Lab at Stanford University and the Audi Electronics Research Lab.


\bibliography{ACC2015}

\end{document}